\documentclass[10pt,twocolumn,letterpaper]{article}

\usepackage{cvpr}
\usepackage{times}
\usepackage{epsfig}
\usepackage{graphicx}
\usepackage{amsmath}
\usepackage{amssymb}

\usepackage{url}            %
\usepackage{booktabs}       %
\usepackage{amsfonts}       %
\usepackage{nicefrac}       %
\usepackage{microtype}      %
\usepackage{textcomp}
\usepackage{xcolor}
\usepackage{multirow}
\usepackage{flushend}
\usepackage{dsfont}

\DeclareMathOperator*{\argmin}{arg\,min}

\usepackage{caption}
\usepackage{subcaption}
\usepackage{paralist}
\usepackage{array}
\usepackage{placeins}
\usepackage{epstopdf}
\usepackage[titletoc,title]{appendix}
\usepackage[T1]{fontenc}
\usepackage{authblk}
\usepackage{diagbox}

\definecolor{mygray}{gray}{0.6}

\usepackage{pifont}%

\usepackage[ruled]{algorithm2e}
\makeatletter
\def\BState{\State\hskip-\ALG@thistlm}
\makeatother

\usepackage{color}
\usepackage{enumitem}
\setitemize{noitemsep,topsep=0pt,parsep=0pt,partopsep=0pt}

\usepackage[pagebackref=true,breaklinks=true,letterpaper=true,colorlinks,bookmarks=false]{hyperref}

\cvprfinalcopy %

\def\cvprPaperID{*} %

\ifcvprfinal\fi
\begin{document}

\title{Learning Representations by Predicting Bags of Visual Words}

\author{
Spyros Gidaris$^{1}$~~~Andrei Bursuc$^{1}$~~~Nikos Komodakis$^{2}$~~~Patrick P{\'e}rez$^{1}$~~~Matthieu Cord$^{1}$\\
\small $^{1}$Valeo.ai~~~~$^{2}$University of Crete
}

\maketitle

\newcommand{\tran}{^\top}
\newcommand{\mtran}{^{-\top}}
\newcommand{\zcol}{\mathbf{0}}
\newcommand{\zrow}{\zcol\tran}

\newcommand{\ind}{\mathbbm{1}}
\newcommand{\expect}{\mathbb{E}}
\newcommand{\nat}{\mathbb{N}}
\newcommand{\zahl}{\mathbb{Z}}
\newcommand{\real}{\mathbb{R}}
\newcommand{\proj}{\mathbb{P}}
\newcommand{\prob}{\mathbf{Pr}}
\newcommand{\softmax}{\operatorname{softmax}}

\newcommand{\defn}{\mathrel{:=}}
\newcommand{\mytexttt}[1]{{\texttt{#1}}}

\newcommand{\vp}{\mathbf{p}}
\newcommand{\vq}{\mathbf{q}}
\newcommand{\vu}{\mathbf{u}}
\newcommand{\vv}{\mathbf{v}}
\newcommand{\vw}{\mathbf{w}}
\newcommand{\vx}{\mathbf{x}}
\newcommand{\vy}{\mathbf{y}}
\newcommand{\vz}{\mathbf{z}}

\newcommand{\phihat}{\hat{\Phi}}

\newcommand{\mv}[1]{\ensuremath{\bm{#1}}} %
\newcommand{\mm}[1]{\ensuremath{\bm{#1}}} %

\newcommand{\Set}[2]{\{\, #1 \mid #2 \, \}}

\newcommand{\few}{\mathrm{few}}
\newcommand{\self}{\mathrm{self}}

\newcommand{\mini}{{MiniImagenet}}
\newcommand{\tmini}{{tiered-MiniImagenet}}
\newcommand{\fscifar}{{CIFAR-FS}}
\newcommand{\VOCseven}{VOC07\xspace}
\newcommand{\ImNet}{ImageNet\xspace}

\newcommand{\resnet}{ResNet\xspace}
\newcommand{\resnetfifty}{ResNet-50\xspace}
\newcommand{\bow}{BoW\xspace}
\newcommand{\bownet}{BoWNet\xspace}

\newcommand{\relu}{\texttt{Relu}\xspace}
\newcommand{\conv}[1]{\mytexttt{conv#1}\xspace}
\newcommand{\pretext}{pretext\xspace}

\makeatletter
\DeclareRobustCommand\onedot{\futurelet\@let@token\@onedot}
\def\@onedot{\ifx\@let@token.\else.\null\fi\xspace}
\def\eg{\emph{e.g}\onedot} \def\Eg{\emph{E.g}\onedot}
\def\ie{\emph{i.e}\onedot} \def\Ie{\emph{I.e}\onedot}
\def\cf{\emph{cf}\onedot} \def\Cf{\emph{Cf}\onedot}
\def\etc{\emph{etc}\onedot} \def\vs{\emph{vs}\onedot}
\def\wrt{w.r.t\onedot} \def\dof{d.o.f\onedot}
\def\etal{\emph{et al}\onedot}
\makeatother
\newcommand{\secref}[1]{\S\ref{#1}}
\begin{abstract}
Self-supervised representation learning targets to learn convnet-based image representations from unlabeled data.
Inspired by the success of NLP methods in this area, in this work we propose a self-supervised approach based on spatially dense image descriptions that encode discrete visual concepts, here called visual words.
To build such discrete representations, we quantize the feature maps of a first pre-trained self-supervised convnet, over a k-means based vocabulary.
Then, as a self-supervised task, we train another convnet to predict the histogram of visual words of an image (\ie, its Bag-of-Words representation) given as input a perturbed version of that image.
The proposed task forces the convnet to learn perturbation-invariant and context-aware image features, useful for downstream image understanding tasks.
We extensively evaluate our method and demonstrate very strong empirical results,
\eg, our pre-trained self-supervised representations transfer better on detection task and similarly on classification over classes ``unseen'' during pre-training, when compared to the supervised case.

This also shows that the process of image discretization into visual words can provide the basis for very powerful self-supervised approaches in the image domain, thus allowing further connections to be made to related methods from the NLP domain that have been extremely successful so far. 
\end{abstract}

\section{Introduction}

\begin{figure*}[!ht]
\renewcommand{\captionfont}{\small}
\centering
\includegraphics[width=0.9\linewidth]{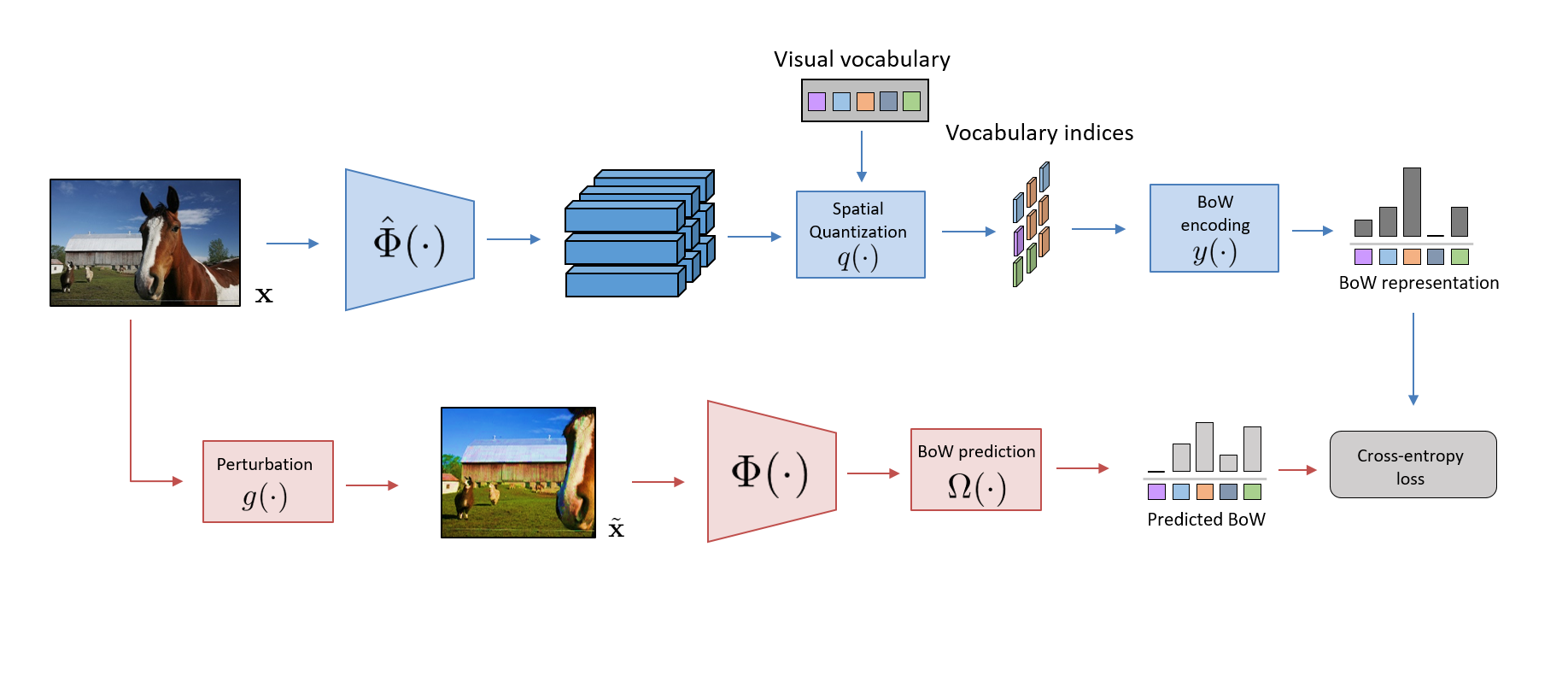}
\vspace{-30pt}
\caption{\textbf{Learning representations through prediction of Bags of Visual Words.} We first train a feature extractor $\phihat(\cdot)$ for a self-supervised task, \eg rotation prediction. Then we compute a visual vocabulary from feature vectors computed from $\phihat$ feature maps and compute the corresponding image level BoW vectors. These BoW vectors will serve as ground truth for the next stage. In the second stage we perturb images with $g(\cdot)$ and send them as input to a second network $\Phi(\cdot)$. The BoW prediction module $\Omega(\cdot)$ processes $\Phi(\cdot)$ feature maps to predict BoW vectors corresponding to the original non-perturbed images. Both $\Phi(\cdot)$ and $\Omega(\cdot)$ are trained jointly with cross-entropy loss. The feature extractor $\Phi(\cdot)$ is further used for downstream tasks. }
\label{fig:bownet_pipeline}
\vspace{-12pt}
\end{figure*}

The goal of our work is to learn convolutional neural network~\cite{lecun1998gradient} (convnet) based representations without human supervision.
One promising approach towards this goal is the so-called self-supervised representation learning \cite{doersch2015unsupervised,zhang2016colorful,larsson2016learning,noroozi2016unsupervised,gidaris2018unsupervised,pathak2016context},
which advocates to train the convnet with an annotation-free pretext task defined using only the information available within an image, \eg, predicting the relative location of two image patches~\cite{doersch2015unsupervised}.
Pre-training on such a pretext task enables the convnet to learn representations that are useful for other vision tasks of actual interest, such as image classification or object detection.
Moreover, recent work has shown that self-supervision can be beneficial to many other learning problems~\cite{su2019does,gidaris2019boosting,zhai2019s,henaff2019data,chen2018self,chen2018self,hendrycks2019using}, such as few-shot~\cite{su2019does,gidaris2019boosting} and semi-supervised~\cite{zhai2019s,henaff2019data} learning, or training generative adversarial networks~\cite{chen2018self}.

A question that still remains open is what type of self-supervision we should use.
Among the variety of the proposed learning tasks, %
many follow the general paradigm of first perturbing an image or removing some part/aspect of the image 
and then training the convnet to reconstruct the original image or the dropped part (\eg, color channel, image region).
Popular examples are Denoising AutoEncoders~\cite{vincent2008extracting}, Image Colorization~\cite{zhang2016colorful,larsson2016learning}, Split-Brain architectures~\cite{zhang2017split}, and Image In-painting~\cite{pathak2016context}. 
However, predicting such low-level image information can be a difficult task to solve, 
and does not necessarily force the convnet to acquire image understanding ``skills'', which is what we ultimately want to achieve.
As a result, such reconstruction-based methods have not been very successful so far. 
In contrast, in Natural Language Processing (NLP), similar self-supervised methods, such as predicting the missing words of a sentence (e.g., BERT~\cite{devlin2018bert} and ROBERTA~\cite{liu2019roberta}), 
have proven much more successful at learning strong language representations. 
The difference of those NLP methods with their computer vision counterparts is that (1) words undoubtedly represent more high-level semantic concepts than raw image pixels. Also, (2) words are defined in a discrete space while images in a continuous one where, without changing the depicted content, small pixel perturbations 
can significantly alter the target of a reconstruction task.

\vspace{-5pt}

\paragraph{Spatially dense image quantization into visual words.}
Inspired by the above NLP methods, in this work we propose for self-supervised learning in the image domain to use tasks that aim at predicting/reconstructing targets that encode \emph{discrete visual concepts} as opposed, \eg, to (low-level) pixel information.
To build such discrete targets, we first take an existing self-supervised method (\eg, rotation prediction~\cite{gidaris2018unsupervised}) and use it to train an initial convnet, which can learn feature representations that capture mid-to-higher-level image features. 
Then, for each image, we densely quantize its convnet-based feature map using a k-means-based vocabulary.\footnote{Here, by dense quantization, we refer to the fact that each spatial location of the feature map is quantized separately.}
This results in a spatially dense image description based on discrete codes (\ie, k-means cluster assignments), called visual words hereafter.
Such a discrete image representation opens the door to easily adapting self-supervised methods from the NLP community to the image domain.
For instance, in this case, one could very well train a BERT-like architecture that, given as input a subset of the patches in an image, predicts the visual words of the missing patches. Although self-supervised methods of this type are definitely something that we plan to explore as future work, in this paper we aim to go one step further and develop (based on the above discrete visual representations) self-supervised tasks that furthermore
allow using standard convolutional architectures that are commonly used (and optimized) for the image domain we are interested in.
But how should we go about defining such a self-supervised task?

\vspace{-5pt}

\paragraph{Learning by ``reconstructing'' bags of visual words.}
To this end, we take inspiration from the so-called Bag-of-Words~\cite{yang2007evaluating} (BoW) model in computer vision
and propose using as self-supervised task one where we wish (to train a convnet) to predict the histogram of visual words of an image (also known as its BoW representation) when given as input a perturbed version of that image.
This type of BoW representations have been very powerful image models, 
and as such have been extensively used in the past in several computer vision problems (including, \eg, image retrieval, object recognition, and object detection).
Interestingly, there is recent empirical evidence that even modern state-of-the-art convnets for image classification exhibit similar behavior to BoW models~\cite{brendel2019approximating}.
By using the above BoW prediction task 
in the context of self-supervised learning, one important benefit 
is that it is no longer required to enhance a typical convnet architecture for images (\eg, \resnetfifty) with extra network components, such as multiple stacks of attention modules as in~\cite{trinh2019selfie} or PixelCNN-like autoregressors as in~\cite{oord2018representation}, that can make the overall architecture computationally intensive.
Furthermore, due to its simplicity, it can be easily incorporated into other types of learning problems (\eg, few-shot learning, semi-supervised learning, or unsupervised domain adaptation), thus allowing to further improve performance for these problems %
which is an additional advantage.

Concerning the perturbed image (that is used as input to the BoW prediction task),
it is generated by applying a set of (commonly used) augmentation techniques such as random cropping, color jittering, or geometric transformations.
Therefore, to solve the task of ``reconstructing'' the BoW histogram of the original image, the convnet must learn to detect visual cues that remain constant (\ie, invariant) to the applied perturbations. Moreover, since the perturbed image can often be only a small part of the original one (due to the cropping transformation), the convnet is also forced to infer the context of the missing input, \ie, the visual words of the missing image regions.
This 
encourages learning of perturbation-invariant and context-aware image features, which, as such, are more likely to encode higher-level semantic visual concepts.
Overall, as we show in the experimental results, this has as a result that the proposed self-supervised method learns representations that transfer significantly better to downstream vision tasks than the representations of the initial convnet.
As a last point, we note that the above process of defining a convnet-based BoW model and then training another convnet to predict it, can be applied iteratively, which can lead to even better representations.

\vspace{-5pt}

\paragraph{Contributions.}To summarize, the contributions of our work are:
\textbf{(1)} We propose the use of discrete visual word representations for self-supervised learning in the image domain.
\textbf{(2)} In this context, we propose a novel method for self-supervised representation learning (Fig.~\ref{fig:bownet_pipeline}). Rather than predicting/reconstructing image-pixel-level information, it uses a first self-supervised pre-trained convnet to densely discretize an image to a set of visual words and then trains a second convnet to predict a reduced Bag-of-Words representation of the image given as input perturbed versions of it.
\textbf{(3)} We extensively evaluate our method and we demonstrate that it manages to learn high-quality convnet-based image representations, which are significantly superior to those of the first convnet.
Furthermore, \emph{our ImageNet-trained self-supervised \resnetfifty representations, when compared to the ImageNet-trained supervised ones,
achieve better VOC07+12 detection performance and comparable Places205 classification accuracy, \ie, better generalization on the detection task and similar generalization on the Places205 classes which are ``unseen'' during self-supervised training.}
\textbf{(4)} 
The simple design of our method allows someone to easily use it on many other learning problems where self-supervision has been shown to be beneficial.

\section{Approach}

Our goal 
is to learn in an unsupervised way a feature extractor or convnet model $\Phi(\cdot)$ parameterized by $\theta$
that, given an image $\vx$, produces a ``good'' image representation $\Phi(\vx)$. By ``good'' we mean a representation that would be useful for other vision tasks of interest, 
\eg image classification, object detection.
To this end, we assume that we have available a large set of unlabeled images $X$
on which we will train our model.
We also assume that we have available an initial self-supervised pre-trained convnet $\hat{\Phi}(\cdot)$.
We can easily learn such a model by employing one of the available self-supervised tasks.
Here, 
except otherwise stated, we use RotNet~\cite{gidaris2018unsupervised} (which is based on the self-supervised task of image rotation prediction) as it is easy to implement and, at the same time, has been shown to achieve strong results in self-supervised representation learning~\cite{kolesnikov2019revisiting}.

To achieve our goal,  we leverage the initial model $\hat{\Phi}(\cdot)$ to create spatially dense descriptions based on visual words.
Then, we 
aggregate those descriptions into BoW representations and train the model $\Phi(\cdot)$ to ``reconstruct'' the BoW of an image $\vx$ given as input a perturbed version of it. 
Note that the 
model $\hat{\Phi}(\cdot)$ remains frozen during the training of the new model $\Phi(\cdot)$. 
Also, after training $\Phi(\cdot)$, we can set $\hat{\Phi}(\cdot) \gets \Phi(\cdot)$ and repeat the training process.

\subsection{Building spatially dense discrete descriptions}

Given a training image $\vx$, the first step for our method is to create a spatially dense visual words-based description $q(\vx)$ using the pre-trained convnet $\hat{\Phi}(\cdot)$.
Specifically, let $\hat{\Phi}(\vx)$ be a feature map (with $\hat{c}$ channels and $\hat{h} \times \hat{w}$ spatial size) produced by $\hat{\Phi}(\cdot)$ for input $\vx$, and $\hat{\Phi}^{u}(\vx)$ the $\hat{c}$-dimensional feature vector at the location $u \in \{1, \cdots, U\}$ of this feature map, where $U = \hat{h} \cdot \hat{w}$. 
To generate the 
description $q(\vx) = [q^1(\vx), \dots, q^{U}(\vx)]$, we densely quantize $\hat{\Phi}(\vx)$ using a predefined vocabulary $V = [\vv_1, ..., \vv_{K}]$ of $\hat{c}$-dimensional visual word embeddings, where $K$ is the vocabulary size.
In detail, for each position $u$, we assign the corresponding feature vector $\hat{\Phi}^{u}(\vx)$ to its closest (in terms of squared Euclidean distance) visual word embedding $q^{u}(\vx)$:
\begin{equation}
    q^{u}(\vx) = \mathop{\argmin_{k=1 \dots K}}\|\hat{\Phi}^{u}(\vx) - \vv_k\|^2_2.
\label{eq:quantization}
\end{equation}
The vocabulary $V$ is learned by applying 
the k-means algorithm with $K$ clusters to a set of feature maps %
extracted from the dataset $X$, \ie, by optimizing the following objective:
\begin{equation}
\mathop{\min_{V}}\sum_{\vx \in X} \sum_{u}\Big[\min_{k=1,\dots,K}{\|\hat{\Phi}_{p}^{u}(\vx) - \vv_k\|^2_2} \Big],
\label{eq:kmeans}
\end{equation}
where the visual word embedding $\vv_k$ is the centroid of the $k$-th cluster.

\subsection{Generating Bag-of-Words representations}  \label{sec:create_bow}

Having generated the discrete description $q(\vx)$ of image $\vx$, the next step is to create its BoW representation, denoted by $y(\vx)$. 
This is a
$K$-dimensional vector whose $k$-th element $y^{k}(\vx)$ either encodes the number of times the $k$-th visual word appears in image $\vx$,
\begin{equation}
y^k(\vx) = \sum_{u=1, \dots, U} \mathds{1}[q^u(\vx) = k],
\label{eq:bow_count}
\end{equation}
or indicates if the $k$-th visual word appears in image $\vx$,
\begin{equation}
y^k(\vx) = \max_{u=1, \dots, U} \mathds{1}[q^u(\vx) = k],
\label{eq:bow_bin}
\end{equation}
where $\mathds{1}[\cdot]$ is the indicator operator.\footnote{In our experiments we use the binary version (\ref{eq:bow_bin})~\cite{sivic2006video, jegou2009packing} for ImageNet and the histogram version (\ref{eq:bow_count}) for CIFAR-100 and MiniImageNet.} 
Furthermore, to convert $y(\vx)$ into a probability distribution over visual words, we $L_1$-normalize it, \ie, we set $y^k(\vx) = \frac{y^k(\vx)}{\sum_{m} y^{m}(\vx)}$.
The resulting $y(\vx)$ can thus be perceived as a soft categorical label of $\vx$ for the $K$ visual words.
Note that, although $K$ might be very large, the BoW representation $y(\vx)$ is actually quite sparse as it has at most $U$ non-zero elements.

\subsection{Learning to ``reconstruct'' BoW} \label{sec:predict_bow}
Based on the above BoW representation, we propose the following self-supervised task:
given an image $\vx$, 
we first apply to it a perturbation operator $g(\cdot)$, to get the perturbed image $\tilde{\vx} = g(\vx)$, and then train the model to predict/``reconstruct'' the BoW representation $y(\vx)$ of the original unperturbed image $\vx$ from $\tilde{\vx}$.
This, in turn, means that we want to predict the BoW representation $y(\vx)$ from the feature vector $\Phi(\tilde{\vx})$ (where hereafter we assume that $\Phi(\tilde{\vx})\in \mathbb{R}^c$, \ie, the feature representation produced by model $\Phi(\cdot)$ is $c$-dimensional).\footnote{\Eg, in the case of ResNet50, $\Phi(\vx)$ corresponds to the $2048$-dimensional feature vector (\ie, $c=2048$) produced from the global average pooling layer that follows the last block of residual layers.}
To this end, we define a 
prediction layer $\Omega(\cdot)$ that gets $\Phi(\tilde{\vx})$ as input and outputs a $K$-dimensional softmax distribution over the $K$ visual words of the BoW representation.
More precisely, the prediction layer is implemented with a linear-plus-softmax layer:
\begin{equation}
    \Omega^{k}(\Phi(\tilde{\vx})) = 
    \mathop{\softmax_{k}}\big[\gamma \Phi(\tilde{\vx})^T \bar{\vw}_m \big]_{m \in [1, \cdots, K]},
\label{eq:predictor}
\end{equation}
where $\Omega^{k}(\Phi(\tilde{\vx}))$ is the softmax probability for the $k$-th visual word, and $W = [\vw_1, \cdots, \vw_K]$ are the $K$ $c$-dimensional weight vectors (one per visual word) of the linear layer. 
Notice that, instead of directly applying the weights vectors $W$ to the feature vector $\Phi(\tilde{\vx})$, we use their $L_2$-normalized versions $\bar{\vw}_k = \vw_k / \|\vw_k\|_2$, and apply a unique learnable magnitude $\gamma$ for all the weight vectors ($\gamma$ is a scalar value).
The reason for this reparametrization of the linear layer is because the
distribution of visual words in the dataset (\ie, how often, or in how many dataset-images, a visual word appears) tends to be unbalanced and, so, without such a reparametrization the network would attempt to make the magnitude of each weight vector proportional to the frequency of its corresponding visual word (thus basically always favoring the most frequently occurring words). In our experiments, the above reparametrization has led to significant improvements in the quality of the learned representations. 

\vspace{-5pt}

\paragraph{Self-supervised training objective.}
The training loss that we minimize for learning the convnet model $\Phi(\cdot)$ is the expected cross-entropy loss between the predicted softmax distribution $\Omega(\Phi(\tilde{\vx}))$ and the BoW distribution $y(\vx)$:
\begin{equation} \label{eq:cos_classifier_loss}
L(\theta, W, \gamma; X) = \mathop{\expect}_{\vx \sim X} \Big[\mathrm{loss}(\Omega(\Phi\big(\tilde{\vx})),y(\vx)\big)\Big] \textrm{,}
\end{equation}
where $\mathrm{loss}(\alpha,\beta) = -\sum_{k=1}^K \beta^k \log \alpha^k$ is the cross-entropy loss for the discrete distributions $\alpha=(\alpha^k)$ and $\beta=(\beta^k)$, 
$\theta$ are the learnable parameters of $\Phi(\cdot)$,
$(W, \gamma)$ are the learnable parameters of $\Omega(\cdot)$,
and $\tilde{\vx} = g(\vx)$.

\vspace{-5pt}

\paragraph{Image perturbations.} 
The perturbation operator $g(\cdot)$ that we use consists of 
(a) color jittering (\ie, random changes of the brightness, contrast, saturation, and hue of an image) 
(b) converting the image to grayscale with probability $p$,
(c) random image cropping,
(d) scale or aspect ratio distortions, and
(e) horizontal flips.
The role served by such an operator is two-fold: to solve the BoW ``reconstruction'' task after such aggressive perturbations, the convnet must learn image features that (1) are robust w.r.t. the applied perturbations and at the same time (2) allow predicting the visual words of the original image, even for image regions that are not visible to the convnet due to cropping.
To further push towards this direction, we also incorporate the CutMix~\cite{yun2019cutmix} augmentation technique into our self-supervised method. 
According to CutMix, given two images $\tilde{\vx}_A = g(\vx_A)$ and $\tilde{\vx}_B = g(\vx_B)$, we generate a new synthetic one $\tilde{\vx}_S$ by replacing a patch of the first image $\tilde{\vx}_A$ with one from the second image $\tilde{\vx}_B$. The position and size of the patch is randomly sampled from a uniform distribution. 
The BoW representation that is used as a reconstruction target for this synthetic image is the convex combination of the BoW targets of the two images, $\lambda y(\vx_A) + (1-\lambda) y(\vx_B)$, where $1-\lambda$ is the patch-over-image area ratio. Hence, with CutMix we force the convnet to infer both (a) the visual words that belong on the patch that was removed from the first image $\tilde{\vx}_A$, and (b) the visual words that belong on the image area that surrounds the patch that was copied from second image $\tilde{\vx}_B$.

\vspace{-5pt}

\paragraph{Model initialization and iterated training.} 
We note that the model $\hat{\Phi}(\cdot)$ is used only for building BoW representations and not for initializing the parameters of the $\Phi(\cdot)$ model, i.e., 
$\Phi(\cdot)$ is randomly initialized before 
training.
Also, as already mentioned, we can apply our self-supervised method iteratively, using each time the previously trained model $\hat{\Phi}(\cdot)$ for creating the BoW representation.
We also note, however, that this is not necessary for learning ``good'' representations; the model learned from the first iteration already achieves very strong results.
As a result, only a few more iterations (\eg, one or two) might be applied after that.

\section{Related Work}

\paragraph{Bag-of-Words.} BoW is a popular method for text document representation, which has been adopted and heavily used in computer vision~\cite{sivic2006video, csurka2004visual}. For visual content, BoW conveniently encapsulates image statistics from hundreds of local features~\cite{lowe2004distinctive} into vector representations. BoW have been studied extensively and leveraged in numerous tasks, while multiple extensions~\cite{perronnin2007fisher, jegou2010aggregating} and theoretical interpretations~\cite{tolias2013aggregate} have been proposed. Due to its versatility, BoW has been applied to pre-trained convnets as well to compute image representations from intermediate feature maps~\cite{yue2015exploiting, gong2014multi, mohedano2016bags}, however few works have dealt with the integration of BoW in the training pipeline of a convnet. Among them, NetVLAD~\cite{arandjelovic2016netvlad} mimics the BoW-derived VLAD descriptor by learning a visual vocabulary along with the other layers and soft quantizing activations over this vocabulary.
Our method differs from previous approaches in training with self-supervision and in predicting directly the BoW vector bypassing quantization and aggregation.

\vspace{-5pt}

\paragraph{Self-supervision.} Self-supervised learning is a recent paradigm aiming to learn representations from data by leveraging supervision from various intrinsic data signals without any explicit manual annotations and human supervision.
The representations learned with self-supervision are then further fine-tuned on a downstream task with limited human annotations available. Numerous creative mechanisms for squeezing out information from data in this manner have been proposed in the past few years: predicting the colors of image~\cite{larsson2016learning, zhang2016colorful}, the relative position of shuffled image patches~\cite{doersch2015unsupervised, noroozi2016unsupervised}, the correct order of a set of shuffled video frames~\cite{misra2016shuffle}, the correct association between an image and a sound~\cite{arandjelovic2017look}, and many other methods~\cite{vondrick2018tracking, zhou2017unsupervised, lee2017unsupervised, wei2018learning}. 

\vspace{-5pt}

\paragraph{Learning to reconstruct.} Multiple self-supervised methods are formulated as reconstruction problems~\cite{vincent2008extracting, mikolov2013efficient, kiros2015skip, zhou2017unsupervised, godard2017unsupervised, zhang2017split, pathak2016context, zhang2016colorful, alayrac2019visual, pillai2019superdepth}. The information to be reconstructed can be provided by a different view~\cite{godard2017unsupervised, zhou2017unsupervised, pillai2019superdepth} or sensor~\cite{eigen2014depth}. When no such complementary information is available, the current data can be perturbed and the task of the model is now to reconstruct the original input. Denoising an input image back to its original state~\cite{vincent2008extracting}, inpainting an image patch that has been removed from a scene~\cite{pathak2016context} 
, reconstructing images that have been overlayed~\cite{alayrac2019visual} are some of the many methods of reconstruction from perturbation. While such approaches display impressive results for the task hand, it remains unclear how much structure they can encapsulate beyond the reconstruction of visual patterns~\cite{dijk2019neural}.
Similar ideas have been initially proposed in NLP where missing words~\cite{mikolov2013efficient} or sentences~\cite{kiros2015skip} must be reconstructed.
Another line of research employs perturbations for guiding the model towards assigning the same label to both perturbed and original content~\cite{dosovitskiy2015discriminative, bachman2019learning}.
Our method also deals with perturbed inputs, however instead of reconstructing the input, we train it to reconstruct the BoW vector of the clean input. This enables learning non-trivial correlations between local visual patterns in the image. 

\vspace{-5pt}

\paragraph{VQ-VAE.}
These works~\cite{oord2017neural, razavi2019generating} explore the learning of spatially-dense discrete representations with unsupervised generative models with goal of image generation. Instead,  we focus on exploiting discrete image representations in the context of self-supervised image representation learning.

\subsection{Discussion}

\noindent \textbf{Relation to clustering-based representation learning methods}~\cite{caron2019unsupervised,caron2018deep,asano2019selflabelling}.
Our work presents similarities to the Deep(er) Clustering approach~\cite{caron2019unsupervised,caron2018deep}.
The latter alternates between k-means clustering the images based on their convnet features and using the cluster assignments as image labels for training the convnet.
In our case however, we use the k-means clustering for creating BoW representations instead of global image labels.
The former leads to richer (more complete) image descriptions compared to the latter as it encodes
 \emph{multiple local visual concepts} extracted in a \emph{spatially dense} way.
For example, 
a cluster id is not 
sufficient to describe an image with multiple objects, like the one in Figure~\ref{fig:bownet_pipeline}, while a BoW is better suited for that.
This fundamental difference leads to a profoundly different self-supervised task.
Specifically, in our case the convnet is forced to:
\textbf{(1)} focus on more localized visual patterns and 
\textbf{(2)} learn better contextual reasoning 
(since it must predict the visual words of missing image regions).
\\
\vspace{-5pt}

\noindent \textbf{Relation to contrastive self-supervised learning methods}~\cite{dosovitskiy2015discriminative, bachman2019learning, he2019momentum, misra2019self}. 
Our method bears similarities to recent works exploiting contrastive losses for learning representations that are invariant under strong data augmentation or perturbations
~\cite{dosovitskiy2015discriminative, bachman2019learning, he2019momentum, misra2019self}.
These methods deal with image recognition and the same arguments mentioned above w.r.t.  \cite{caron2019unsupervised,caron2018deep} hold. 
Our contribution departs from this line of approaches allowing our method to be applied to a wider set of visual tasks.
For instance, in autonomous driving, most urban images are similar and differ only by few details,
\eg a pedestrian or a car, making
image recognition under strong perturbation less feasible.
In such cases, leveraging local statistics as done in our method
appears as a more appropriate self-supervised task for learning 
representations.

\section{Experiments and results}

We evaluate our method (BoWNet) on CIFAR-100, MiniImageNet~\cite{vinyals2016matching}, ImageNet~\cite{russakovsky2015imagenet}, Places205~\cite{NIPS2014_5349}, VOC07~\cite{Everingham10} classification, and V0C07+12 detection datasets. 

\subsection{Analysis on CIFAR-100 and MiniImageNet} \label{sec:method_analysis}

\subsubsection{Implementation details}

\textbf{CIFAR-100.} CIFAR-100 consists of $50k$ training images with $32\times32$ resolution.
We train self-supervised WRN-28-10~\cite{zagoruyko2016wide} convnets using those training images.
Specifically, we first train a WRN-28-10 based RotNet~\cite{gidaris2018unsupervised} and build based on that BoW using the feature maps of its last/3rd residual block. 
Then, we train the BoWNet using those BoW representations. 
We use a $K=2048$ vocabulary size.
The prediction head of RotNet consists of an extra residual block (instead of just a linear layer); in our experiments this led the feature extractor to learn better representations (we followed this design choice of RotNet for all the experiments in our paper; we provide more implementation details in~\S\ref{sec:imp_details_rotnet}). 

We train the convnets using stochastic gradient descent (SGD) for $30$ epochs of $2000$ batch iterations and batch size $128$.
The learning rate is initialized with $0.1$ and multiplied by $0.1$ after $12$ and $24$ epochs. 
The weight decay is $5e-4$.

\textbf{MiniImageNet.} Since MiniImageNet is used for evaluating few-shot methods, it has three different splits of classes, train, validation, and test with $64$, $16$, and $20$ classes respectively. Each class has $600$ images with $84 \times 84$ resolution.
We train WRN-28-4 convnets on the $64 \times 600$ images that correspond to the training classes following the same training protocol as for CIFAR-100.

\subsubsection{Evaluation protocols.}

\textbf{CIFAR-100.} To evaluate the learned representations we use two protocols.
\textbf{(1)} The first is to freeze the learned representations (which in case of WRN-28-10 is a $640$-dimensional vector) and train on top of them a $100$-way linear classifier for the CIFAR-100 classification task. 
We use the linear classifier accuracy as an evaluation metric.
The linear classifier is trained with SGD for $60$ epochs using a learning rate of $0.1$ that is multiplied by $0.3$ every $10$ epochs.
The weight decay is $0.001$. 
\textbf{(2)}
For the second protocol we use a few-shot episodic setting~\cite{vinyals2016matching} similar to what is proposed on~\cite{gidaris2019boosting}.
Specifically, we choose $20$ classes from CIFAR-100 and run with them multiple ($2000$) episodes of $5$-way few-shot classification tasks.
Essentially, at each episode we randomly sample $5$ classes from the $20$ ones and then $n$ training examples and $m=15$ test examples per class (both randomly sampled from the test images of CIFAR-100). 
For $n$ we use $1$, $5$, $10$, and $50$ examples ($1$-shot, $5$-shot, $10$-shot, and $50$-shot settings respectively).
To classify the $m$ examples we use a cosine distance Prototypical-Networks~\cite{snell2017prototypical} classifier that is applied on top of the frozen representations.
We report the mean accuracy over the $2000$ episodes. 
The purpose of this metric is to analyze the ability of the representations to be used for learning with few training examples.
More details about this protocol are provided in~\S\ref{sec:imp_details_fewshot}.

\textbf{MiniImageNet.} 
We use the same two protocols as in CIFAR-100.
\textbf{(1)} The first is to train $64$-way linear classifiers on the task of recognizing the $64$ training classes of MiniImageNet. Here, we use the same hyper-parameters as for CIFAR-100.
\textbf{(2)}
The second protocol is to use the frozen representations for episodic few-shot classification~\cite{vinyals2016matching}.
The main difference with CIFAR-100 is that here we evaluate using the test classes of MiniImageNet, which were not 
part of the training set of the self-supervised models.
Therefore, with this evaluation we analyze the ability of the representations to be used for learning with few training examples and for ``unseen" classes during training.
For comparison with this protocol we provide results of the supervised Cosine Classifier (CC) few-shot model~\cite{gidaris2018dynamic,qi2017learning}.

\subsubsection{Results}
\begin{table}[t!]
\centering
\renewcommand{\figurename}{Table}
\renewcommand{\captionlabelfont}{\bf}
\renewcommand{\captionfont}{\small} 
{\setlength{\extrarowheight}{2pt}\small
{
\begin{tabular}{ l  | c  c  c  c | c  }
\toprule
\multicolumn{1}{l|}{Method} & \multicolumn{1}{c}{$n=$1} & \multicolumn{1}{c}{5} & \multicolumn{1}{c}{10} & \multicolumn{1}{c|}{50} & \multicolumn{1}{c}{Linear}\\
\midrule
\;Supervised~\cite{zagoruyko2016wide} & - & - & - & - & 79.5\\
\midrule
\;RotNet            & 58.3 & 74.8 & 78.3 & 81.9 & 60.3\\
\;Deeper Clustering & 65.9 & 84.6 & 87.9 & 90.8 & 65.4\\
\;\textcolor{mygray}{AMDIM~\cite{bachman2019learning}} & - & - & - & - & \textcolor{mygray}{70.2}\\
\midrule
\;BoWNet            & \textbf{69.1} & 86.3 & 89.2 & 92.4 & 71.5\\  
\;BoWNet $\times2$  & 68.5 & 87.1 & \textbf{90.4} & 93.8 & 74.1\\
\;BoWNet $\times3$  & 68.4 & \textbf{87.2} & \textbf{90.4} & \textbf{93.9} & \textbf{74.5}\\
\midrule
\;BoWNet w/o cutmix & 68.5 & 85.8 & 88.8 & 92.2 & 69.7\\
\;Sp-BoWNet    & 67.7 & 85.8 & 89.2 & 92.3 & 71.3\\
\bottomrule
\end{tabular}}}
\caption{
\textbf{CIFAR-100 linear classifier and few-shot results with WRN-28-10.}
For few-shot we use $n$=1, 5, 10, or 50 examples per class. 
AMDIM uses a higher-capacity custom made architecture.}
\vspace{-12pt}
\label{tab:UnsupervisedCIFAR100Results}
\end{table}

\begin{table}[t!]
\centering
\renewcommand{\figurename}{Table}
\renewcommand{\captionlabelfont}{\bf}
\renewcommand{\captionfont}{\small} 
{\setlength{\extrarowheight}{2pt}\small
{
\begin{tabular}{ l  | c  c  c  c | c  }
\toprule
\multirow{2}{*}{\diagbox{Method}{Classes}} & \multicolumn{4}{c|}{Novel} & \multicolumn{1}{c}{Base}\\
& \multicolumn{1}{c}{$n=$1} & \multicolumn{1}{c}{5} & \multicolumn{1}{c}{10} & \multicolumn{1}{c|}{50} & \multicolumn{1}{c}{Linear}\\
\midrule
\;Supervised CC~\cite{gidaris2018dynamic} & 56.8 & 74.1 & 78.1 & 82.7 & 73.7 \\
\midrule
\;RotNet             & 40.8 & 56.9 & 61.8 & 68.1 & 52.3\\
\;RelLoc~\cite{doersch2015unsupervised}             & 40.2 & 57.1 & 62.6 & 68.8 & 50.4\\
\;Deeper Clustering  & 47.8 & 66.6 & 72.1 & 78.4 & 60.3\\
\midrule
\;BoWNet             & 48.7 & 67.9 & 74.0 & 79.9 & 65.0\\  
\;BoWNet $\times2$   & \textbf{49.1} & 67.6 & 73.6 & 79.9 & 65.6\\
\;BoWNet $\times3$   & 48.6 & \textbf{68.9} & \textbf{75.3} & \textbf{82.5} & \textbf{66.0}\\
\bottomrule
\end{tabular}}}
\caption{
\textbf{MiniImageNet linear classifier and few-shot results with WRN-28-4.}
}
\vspace{-12pt}
\label{tab:UnsupervisedMiniImageNetResults}
\end{table}

\begin{table}[t!]
\centering
\renewcommand{\figurename}{Table}
\renewcommand{\captionlabelfont}{\bf}
\renewcommand{\captionfont}{\small} 
{\setlength{\extrarowheight}{2pt}\small
{
\begin{tabular}{ l  | c  c  c  c | c  }
\toprule
\multirow{2}{*}{\diagbox{Method}{Classes}} & \multicolumn{4}{c|}{Novel} & \multicolumn{1}{c}{Base}\\
& \multicolumn{1}{c}{$n=$1} & \multicolumn{1}{c}{5} & \multicolumn{1}{c}{10} & \multicolumn{1}{c|}{50} & \multicolumn{1}{c}{Linear}\\
\midrule
\;RotNet                & 40.8 & 56.9 & 61.8 & 68.1 & 52.3\\
\;RelLoc                & 40.2 & 57.1 & 62.6 & 68.8 & 50.4\\
\midrule
\;RotNet $\to$ BoWNet  & 48.7 & 67.9 & 74.0 & 79.9 & 65.0\\
\;RelLoc $\to$ BoWNet  & \textbf{51.8} & \textbf{70.7} & \textbf{75.9} & \textbf{81.3} & \textbf{65.2}\\
\;Random $\to$ BoWNet & 42.4 & 62.0 & 68.9 & 78.1 & 61.5\\
\bottomrule
\end{tabular}}}
\caption{
\textbf{MiniImageNet linear classifier and few-shot results with WRN-28-4. Impact of base convnet.}
}
\vspace{-12pt}
\label{tab:UnsupervisedMiniImageNetResultsBaseNet}
\end{table}

In Tables~\ref{tab:UnsupervisedCIFAR100Results} and~\ref{tab:UnsupervisedMiniImageNetResults} we report results 
for our self-supervised method on the CIFAR-100 and MiniImageNet datasets respectively.
By comparing BoWNet with RotNet (that we used for building BoW), we observe that BoWNet improves all the evaluation metrics by at least $10$ percentage points, which is a very large performance improvement.
Applying BoWNet iteratively (entries BoWNet $\times2$ and BoWNet $\times3$) further improves the results (except the $1$-shot accuracy). %
Also, BoWNet outperforms by a large margin the CIFAR-100 linear classification accuracy of the recently proposed AMDIM~\cite{bachman2019learning} method (see Table~\ref{tab:UnsupervisedCIFAR100Results}), which has been shown to achieve very strong results.
Finally, the performance of the BoWNet representations on the MiniImageNet novel classes for the $10$-shot and especially the $50$-shot setting are very close to the that of the supervised CC model (see Table~\ref{tab:UnsupervisedMiniImageNetResults}).

\textbf{Impact of CutMix augmentation.}
In Table~\ref{tab:UnsupervisedCIFAR100Results} we report CIFAR-100 results without CutMix, which confirms that employing CutMix does indeed provide some further improvement on the quality of the learned representations.

\textbf{Spatial-Pyramid BoW~\cite{lazebnik2006beyond}.}
By reducing the visual words descriptions to BoW histograms, 
we remove spatial information from the visual word representations.
To avoid this, one could divide the image into several spatial grids of different resolutions, and then extract a BoW representation from each resulting image patch.
In Table~\ref{tab:UnsupervisedCIFAR100Results} we provide results for such a case (entry Sp-BoWNet).
Specifically, we used 2 levels for the spatial pyramid, one with $2 \times 2$ and one with $1 \times 1$ resolution, giving in total $5$ BoW.
Although one would expect otherwise, we observe that adding more spatial information to the BoW via Spatial-Pyramids-BoW, does not improve the quality of the learned representations.

\textbf{Comparison with Deeper Clustering (DC)~\cite{caron2019unsupervised}.}
To compare our method against DC we implemented it using the same convnet backbone and the same pre-trained RotNet as for BoWNet.
To be sure that we do not disadvantage DC in any way,
we optimized the number of clusters ($512$), the training routine\footnote{Specifically, we use $30$ training epoch (the clusters are updated every $3$ epochs), and a constant learning rate of 0.1 (same as in~\cite{caron2019unsupervised}). Each epoch consists of $2000$ batch iterations with batch size $512$. For simplicity we used one clustering level instead of the two hierarchical levels in~\cite{caron2019unsupervised}.}, 
and applied the same image augmentations (including cutmix) as in BoWNet.
We also boosted DC by combining it with rotation prediction and applied our re-parametrization of the linear prediction layer (see equation (\ref{eq:predictor})), which however did not make any difference in the DC case.
We observe in Tables~\ref{tab:UnsupervisedCIFAR100Results} and~\ref{tab:UnsupervisedMiniImageNetResults} that, although improving upon RotNet, DC has finally a significantly lower performance compared to BoWNet, e.g., several absolute percentage points lower linear classification accuracy, 
which illustrates the advantage of using BoW as targets for self-supervision instead of the single cluster id of an image.

\textbf{Impact of base convnet.}
In Table~\ref{tab:UnsupervisedMiniImageNetResultsBaseNet} we provide MiniImageNet results using RelLoc~\cite{doersch2015unsupervised} as the initial convnet with which we build BoW (base convnet).
RelLoc $\to$ BoWNet achieves equally strong or better results than in the RotNet $\to$ BoWNet case.
We also conducted (preliminary) experiments with a randomly initialized base convnet (entry Random $\to$ BoWNet).
In this case, to learn good representations, 
(a) we used in total 4 training rounds, (b) for the 1st round we built BoW from the 1st residual block of the randomly initialized WRN-28-4 and applied PCA analysis before k-means, (c) for the 2nd round we built BoW from 2nd residual block, and (d) for the remaining 2 rounds we built BoW from 3rd/last residual block.
We observe that with a random base convnet the performance of BoWNet drops. 
However, BoWNet still is significantly better than RotNet and RelLoc.

We provide additional experimental results in~\S\ref{sec:additional_exp_cifar100}.  

\subsection{Self-supervised training on ImageNet} \label{sec:large_scale_experiments}

Here we evaluate BoWNet by training it on the ImageNet dataset that consists of more than $1M$ images coming from $1000$ different classes.
We use the \resnetfifty (v1)~\cite{he2016deep} architecture with $224 \times 224$ input images for implementing the RotNet and BoWNet models.
The BoWNet models are trained with 2 training rounds. 
For each round we use SGD, $140$ training epochs, and a learning rate $0.03$ that is multiplied by $0.1$ after $60$, $100$, and $130$ epochs.
The batch size is $256$ and the weight decay is $1e-4$.
To build BoW we use a vocabulary of $K=20000$ visual words created from the 3rd or 4th residual blocks (aka \texttt{conv4} and \texttt{conv5} layers respectively) of RotNet (for an experimental analysis of those choices see~\S\ref{sec:imagenet_BoW_design}).
We named those two models BowNet \texttt{conv4} and BowNet \texttt{conv5} respectively.

We evaluate the quality of the learned representations on 
ImageNet classification, Places205 classification, VOC07 classification, and VOC07+12 detection tasks.

\begin{table}[t!]
\centering
\renewcommand{\figurename}{Table}
\renewcommand{\captionlabelfont}{\bf}
\renewcommand{\captionfont}{\small} 
{\setlength{\extrarowheight}{2pt}\small
{
\begin{tabular}{ l  |  c   c  }
\toprule
\multicolumn{1}{l|}{Method} &  \texttt{conv4} &  \texttt{conv5}  \\
\midrule
\ImNet Supervised~\cite{goyal2019scaling} & 80.4 & 88.0  \\
\midrule
RotNet$^\ast$ & 64.6 & 62.8 \\
Jigsaw~\cite{goyal2019scaling}    & 64.5 & 57.2 \\
Colorization~\cite{goyal2019scaling}   &  55.6 & 52.3 \\  
\midrule
BoWNet \texttt{conv4} & 73.6 & 79.3\\
BoWNet \texttt{conv5} & 74.3 & 78.4\\
\bottomrule
\end{tabular}}}
\caption{
\textbf{VOC07 image classification results for ResNet-50 Linear SVMs}. $^\ast$: our implementation.}
\vspace{-12pt}
\label{tab:ClassificationVOC07}
\end{table}

\begin{table}[t!]
\centering
\renewcommand{\figurename}{Table}
\renewcommand{\captionlabelfont}{\bf}
\renewcommand{\captionfont}{\small} 
{\setlength{\extrarowheight}{2pt}\small
{
\begin{tabular}{ l  | c c | c c }
\toprule
       & \multicolumn{2}{c|}{ImageNet}   & \multicolumn{2}{c}{Places205}\\    
Method & \texttt{conv*} & \texttt{pool5} & \texttt{conv*} & \texttt{pool5}\\
\midrule
\;Random~\cite{goyal2019scaling}    & 13.7 & -    & 16.6 & -\\
\midrule
\multicolumn{5}{l}{\textbf{\;Supervised methods}}\\
\;ImageNet~\cite{goyal2019scaling}  & 75.5 & -    & 51.5 & -\\
\;ImageNet$^\ast$                   & 76.0 & 76.2 & 52.8 & 52.0\\
\;Places205~\cite{goyal2019scaling} & 58.2 & -    & 62.3 & -\\
\midrule
\multicolumn{5}{l}{\textbf{\;Prior self-supervised methods}}\\
\;RotNet$^\ast$                         & 52.5 & 40.6 & 45.0 & 39.4\\
\;Jigsaw~\cite{goyal2019scaling}        & 45.7 & -    & 41.2 & -\\
\;Colorization~\cite{goyal2019scaling}  & 39.6 & -    & 31.3 & -\\  
\;\textcolor{mygray}{LA$^\dagger$~\cite{zhuang2019local}}  & \textcolor{mygray}{60.2} & - & \textcolor{mygray}{50.1} & -\\
\midrule
\multicolumn{5}{l}{\textbf{\;Concurrent work}}\\
\;MoCo~\cite{he2019momentum}      & - & \underline{60.6} & - & - \\
\;PIRL~\cite{misra2019self}       & \textbf{63.6} & - & 49.8 & - \\
\;\textcolor{mygray}{CMC$^\ddagger$~\cite{tian2019contrastive}} & - & \textcolor{mygray}{64.1} & - & - \\
\midrule
\;BowNet \texttt{conv4}            & \underline{62.5} & \textbf{62.1} & \textbf{50.9} & \textbf{51.1}\\
\;BowNet \texttt{conv5}            & 60.5 & 60.2 & \underline{50.1} & \underline{49.5}\\
\bottomrule
\end{tabular}}}
\caption{
\textbf{\resnetfifty top-1 center-crop linear classification accuracy on ImageNet and Places205.} 
\texttt{pool5} indicates the accuracy for the $2048$-dimensional features produced by the global average pooling layer after \texttt{conv5}.
\texttt{conv*} indicates the accuracy of the best (w.r.t. accuracy) conv. layer of \resnetfifty (for the full results see~\S\ref{sec:full_imnet_places}).
Before applying classifiers on those conv. layers, we resize their feature maps to around $9k$ dimensions (in the same way as in~\cite{goyal2019scaling}).
$^\dagger$: LA~\cite{zhuang2019local} uses 10-crops evaluation.
$^\ddagger$: CMC~\cite{tian2019contrastive} uses two \resnetfifty feature extractor networks.
$^\ast$: our implementation.}
\vspace{-12pt}
\label{tab:ImageNetPlacesResultsCompressed}
\end{table}

\begin{table}[t!]
\centering
\renewcommand{\figurename}{Table}
\renewcommand{\captionlabelfont}{\bf}
\renewcommand{\captionfont}{\small} 
{\setlength{\extrarowheight}{2pt}\small
{
\begin{tabular}{ l  | c  c  c   }
\toprule
\multicolumn{1}{l|}{Pre-training} & \multicolumn{1}{c}{$\text{AP}^{\text{50}}$} & \multicolumn{1}{c}{$\text{AP}^{75}$} & \multicolumn{1}{c}{$\text{AP}^{all}$} \\
\midrule
Supervised$^\ast$ & 80.8 & 58.5 & 53.2 \\
\midrule
Jigsaw$^\dagger$~\cite{goyal2019scaling} & 75.1 & 52.9 & 48.9 \\
PIRL$^\dagger$~\cite{misra2019self} & 80.7 & 59.7 & 54.0 \\
MoCo~\cite{he2019momentum} & \textbf{81.4} & \textbf{61.2} & \underline{55.2} \\

\midrule
\bownet \texttt{conv4} & 80.3 & 60.4 & 55.0 \\            
\bownet \texttt{conv5} & \underline{81.3} & \underline{61.1} & \textbf{55.8} \\     
\bottomrule
\end{tabular}}}
\caption{
\textbf{
Object detection with Faster R-CNN fine-tuned on VOC \texttt{trainval07+12}.}
The detection AP scores ($\text{AP}^{\text{50}}$, $\text{AP}^{\text{75}}$, $\text{AP}^{\text{all}}$) are computed on \texttt{test07}.
All models use ResNet-50 backbone (R50-C4) pre-trained with self-supervision on ImageNet.
BowNet scores are averaged over 3 trials. $^\ast$: our implementation fine-tuned in the same conditions as \bownet. $^\dagger$: BatchNorm layers are frozen and used as affine transformation layers.
}
\vspace{-12pt}
\label{tab:VOCDetectionResults}
\end{table}

\vspace{-5pt}

\paragraph{VOC07 classification results.} %
For this evaluation we use the publicly available code for benchmarking self-supervised methods provided by Goyal \etal~\cite{goyal2019scaling}.
\cite{goyal2019scaling} implements the guidelines of~\cite{owens2016ambient} and trains linear SVMs~\cite{boser1992training} on top of the frozen learned representations using the VOC07 train+val splits for training and the VOC07 test split for testing.
We consider the features of the 3rd (layer \texttt{conv4}) and 4th (layer \texttt{conv5}) residual blocks and provide results in Table~\ref{tab:ClassificationVOC07}.
Again, BoWNet improves the performance of the already strong RotNet by several points.
Furthermore, BoWNet outperforms all prior methods.
Interestingly, \texttt{conv4}-based BoW leads to better classification results for the \texttt{conv5} layer of BoWNet, and \texttt{conv5}-based BoW leads to better classification results for the \texttt{conv4} layer of BoWNet.

\vspace{-5pt}

\paragraph{ImageNet and Places205 classification results.}
Here we evaluate on the $1000$-way ImageNet and $205$-way Places205 classification tasks using linear classifiers on frozen feature representations.
To that end, we follow the guidelines of \cite{goyal2019scaling}: for the ImageNet (Places205) dataset we train linear classifiers using Nesterov SGD for $28$ ($14$) training epochs and a learning rate of $0.01$ that is multiplied by $0.1$ after $10$ ($5$) and $20$ ($10$) epochs.
The batch size is $256$ and the weight decay is $1e-4$.
We report results in Table~\ref{tab:ImageNetPlacesResultsCompressed}.
We observe that BoWNet outperforms all prior self-supervised methods by significant margin.
Furthermore, the accuracy gap on Places205 between our ImageNet-trained BoWNet representations and the ImageNet-trained supervised representations is only $0.9$ points in \texttt{pool5}.
This demonstrates that \emph{our self-supervised representations have almost the same generalization ability to the ``unseen'' 
(during training) Places205 classes as the supervised ones}.
We also compare against the MoCo~\cite{he2019momentum} and  PIRL~\cite{misra2019self} methods that were recently uploaded on arXiv and essentially are concurrent work.
BoWNet outperforms MoCo on ImageNet. 
When compared to PIRL, BoWNet has around $1$ point higher Places205 accuracy but $1$ point lower ImageNet accuracy.

\vspace{-5pt}

\paragraph{VOC detection results.}
Here we evaluate the utility of our self-supervised method on a more complex downstream task: object detection. We follow the setup considered in prior works~\cite{goyal2019scaling, he2019momentum, misra2019self}: Faster R-CNN~\cite{ren2015faster} with a ResNet50  backbone~\cite{he2017mask} (R50-C4 in Detectron2~\cite{wu2019detectron2}). We fine-tune the pre-trained \bownet on \texttt{trainval07+12} and evaluate on \texttt{test07}.
We use the same training schedule as \cite{goyal2019scaling, misra2019self} adapted for 8 GPUs and freeze the first two convolutional blocks. In detail, we use mini-batches of $2$ images per GPU and fine-tune for $25K$ steps with the learning rate dropped by $0.1$ after $17K$ steps. We set the base learning to $0.02$ with a linear warmup~\cite{goyal2017accurate} of $1,000$ steps.
We fine-tune BatchNorm layers~\cite{ioffe2015batch} (synchronizing across GPUs) and use BatchNorm on newly added layers specific to this task.\footnote{He \etal \cite{he2019momentum} point out that features produced by self-supervised training can display different distributions compared to supervised ones and suggest using feature normalization to alleviate this problem.} 

We compare \bownet \texttt{conv4} and \bownet \texttt{conv5} against both classic and recent self-supervised methods and report results in Table~\ref{tab:VOCDetectionResults}. Both \bownet variants exhibit strong performance. Differently from previous benchmarks, the \texttt{conv5} is clearly better than \texttt{conv4} on all metrics. This might be due to the fact that  here we fine-tune multiple layers and depth plays a more significant role. 
\emph{Interestingly, \bownet outperforms the supervised ImageNet pre-trained model}, which is fine-tuned in the same conditions as \bownet.
So, \emph{our self-supervised representations generalize better to the VOC detection task than the supervised ones}.
This result is in line with concurrent works \cite{he2019momentum, misra2019self} and underpins the utility of such methods in efficiently squeezing out information from data without using labels.  

\section{Conclusion}

In this work we propose BoWNet, a novel method for representation learning that employs spatial dense descriptions based on visual words as targets for self-supervised training. 
The labels for training BoWNet are provided by a standard self-supervised model. The reconstruction of the BoW vectors from perturbed images along with the discretization of the output space into visual words, enable a more discriminative learning of the local visual patterns in an image. Interestingly, although BoWNet is trained over features learned without label supervision, not only it achieves strong performances, but it also manages to outperform the initial model. This finding along with the discretization of the feature space (into visual words) open additional perspectives and bridges to NLP self-supervised methods that have greatly benefited from this type of approaches in the past few years.

\vspace{-5pt}
\paragraph{Acknowledgements.} \small We would like to thank Gabriel de Marmiesse for his invaluable support during the experimental implementation and analysis of this work.

{\small
\bibliographystyle{ieee_fullname}
\bibliography{egbib}
}

\clearpage
\appendix
\section*{Appendix}

\begin{figure}[!ht]
\renewcommand{\captionfont}{\small}
\centering
\includegraphics[width=0.9\linewidth]{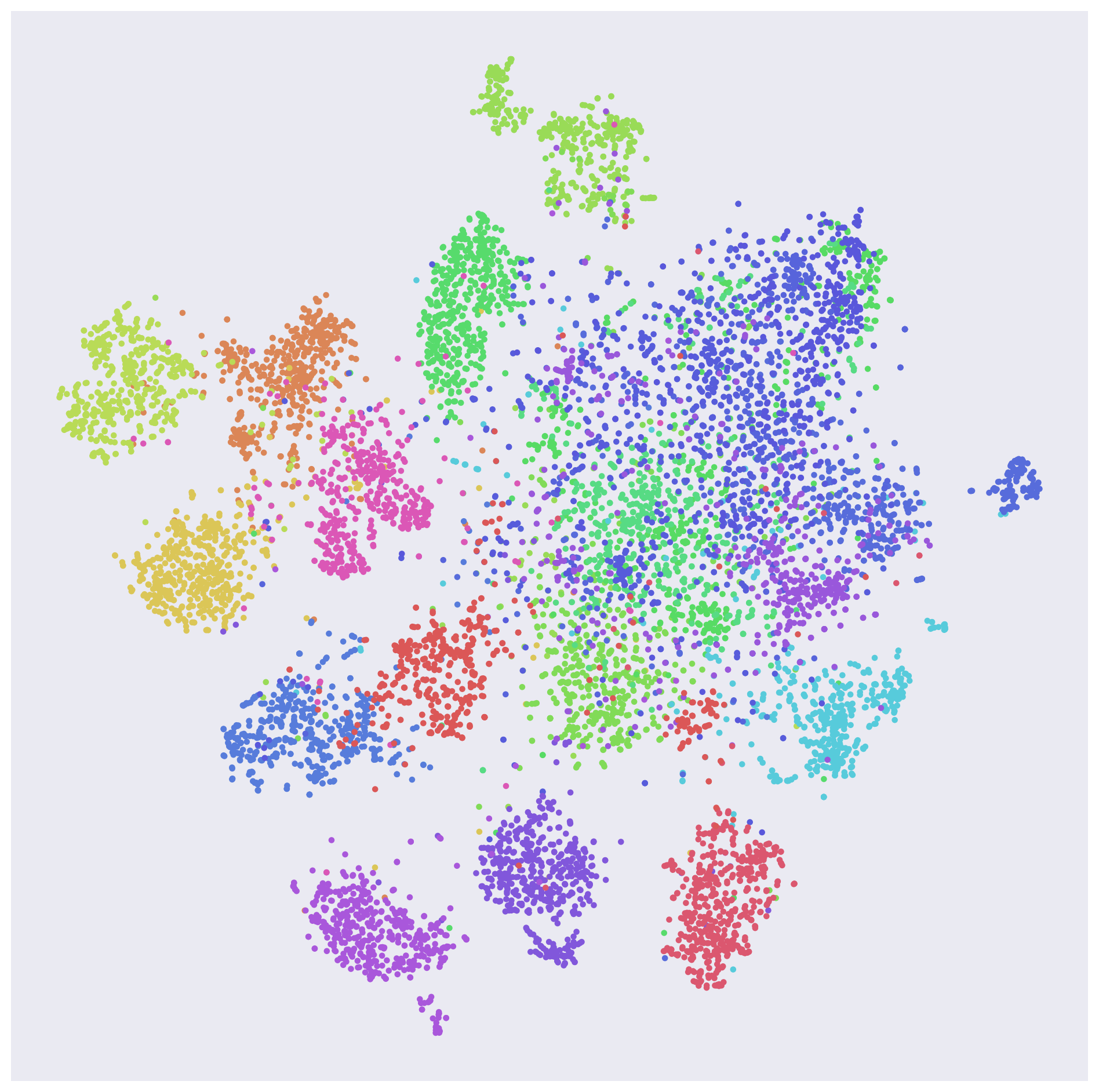}
\caption{\textbf{t-SNE~\cite{maaten2008visualizing} scatter plot of the learnt self-supervised features on CIFAR-100.} 
Each data point in the t-SNE scatter plot corresponds to the self-supervised feature representation of an image from CIFAR-100 and is colored according to the class that this image belongs to.
To reduce clutter, we visualize the features extracted from the images of 20 randomly selected classes of CIFAR-100.
}
\label{fig:tsne}
\end{figure}
\begin{figure*}
    \centering
    \begin{tabular}{cc}
    \includegraphics[width=0.45\linewidth]{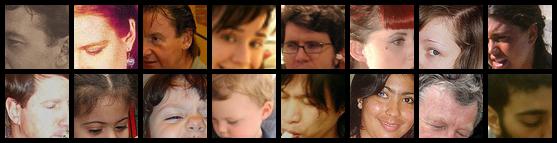} & \includegraphics[width=0.45\linewidth]{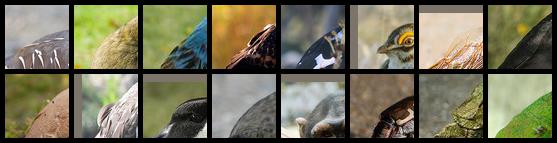} \\
    \includegraphics[width=0.45\linewidth]{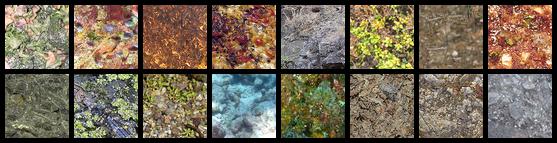} & \includegraphics[width=0.45\linewidth]{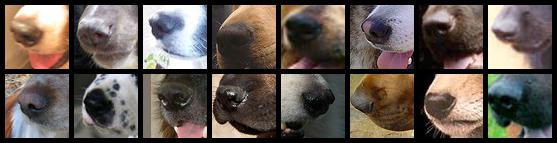} \\  
    \includegraphics[width=0.45\linewidth]{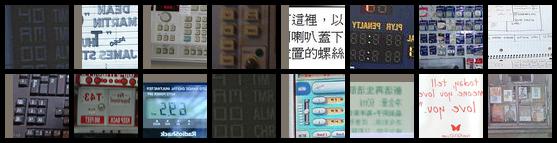} & \includegraphics[width=0.45\linewidth]{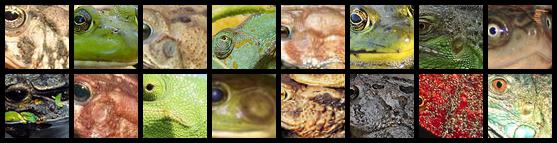} \\
    \includegraphics[width=0.45\linewidth]{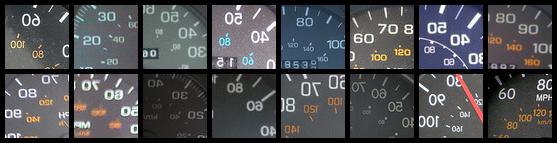} & \includegraphics[width=0.45\linewidth]{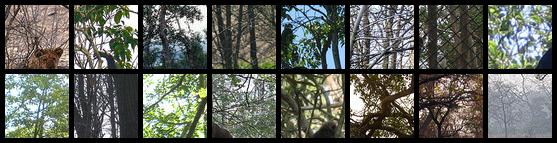} \\  
    \includegraphics[width=0.45\linewidth]{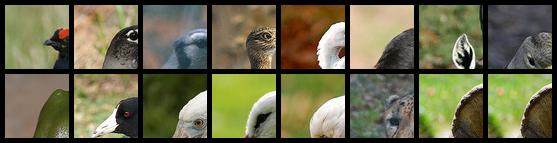} & \includegraphics[width=0.45\linewidth]{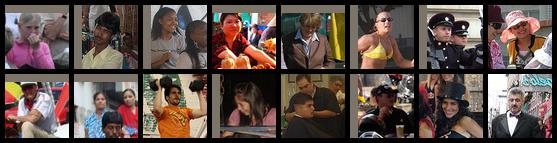} \\ 
    \includegraphics[width=0.45\linewidth]{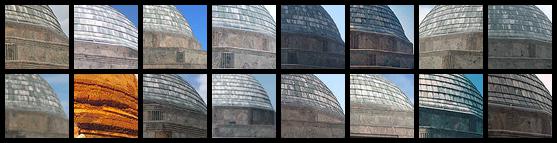} & \includegraphics[width=0.45\linewidth]{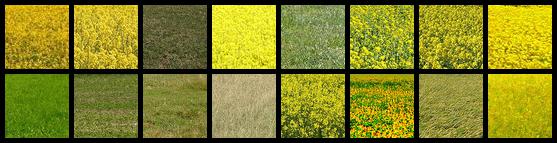} \\ 
    \includegraphics[width=0.45\linewidth]{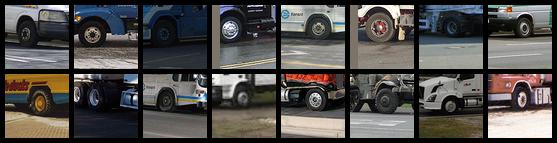} & \includegraphics[width=0.45\linewidth]{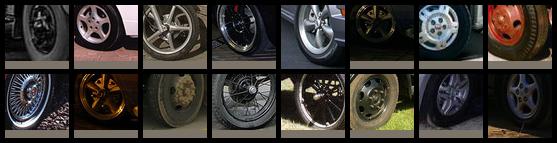} \\  
    \includegraphics[width=0.45\linewidth]{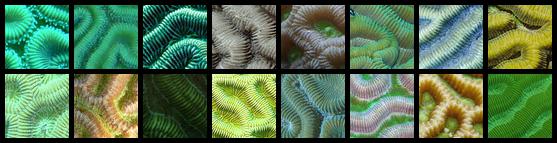} & \includegraphics[width=0.45\linewidth]{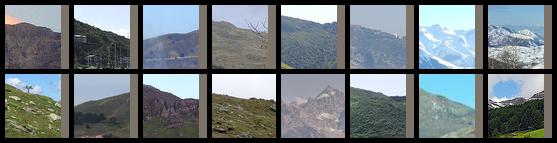} \\
    \includegraphics[width=0.45\linewidth]{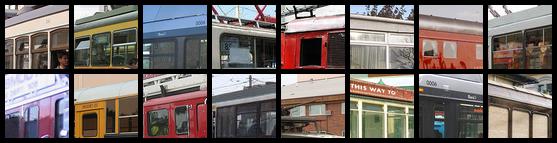} & \includegraphics[width=0.45\linewidth]{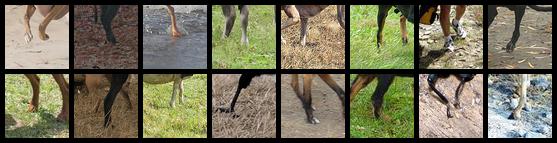} \\
    \includegraphics[width=0.45\linewidth]{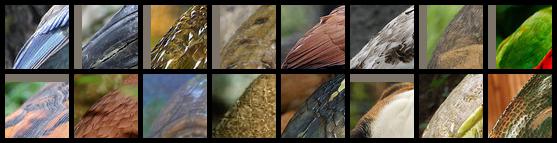} & \includegraphics[width=0.45\linewidth]{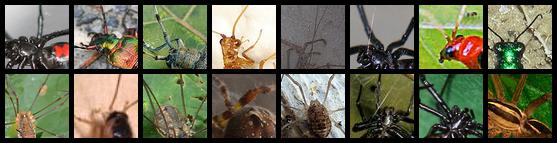} \\     
    \end{tabular}
    \caption{\textbf{Examples of visual word cluster members.} The clusters are created by applying k-means on the feature maps produced from the \texttt{conv4} layer of the ResNet50.
    For each visual word cluster we depict the 16 patch members with the smallest Euclidean distance to the visual word cluster centroid.}
    \label{fig:visual_words}
\end{figure*}

\section{Visualizations}

\subsection{Visualizing the word clusters}
In Figure~\ref{fig:visual_words} we illustrate visual words used for training our self-supervised method on ImageNet. 
Since we discover visual words using k-means, 
to visualize a visual word we depict the 16 patches with the smallest Euclidean distance to the visual word cluster centroid.
As can be noticed, visual words encode mid-to-higher-level visual concepts.

\subsection{t-SNE scatter plots of the learnt self-supervised features}
In Figure~\ref{fig:tsne} we visualize the t-SNE~\cite{maaten2008visualizing} scatter plot of the self-supervised features
obtained when applying our method to the CIFAR-100 dataset.
For visualizations purposes, we only plot the features corresponding to images that belong to 20 (randomly selected) classes out of the 100 classes of CIFAR-100.
As can be clearly observed, the learnt features form class-specific clusters, which indicates that they capture semantic image information.
\begin{table}[t!]
\centering
\renewcommand{\figurename}{Table}
\renewcommand{\captionlabelfont}{\bf}
\renewcommand{\captionfont}{\small} 
{\setlength{\extrarowheight}{2pt}\small
{
\begin{tabular}{ l  |  c c }
\toprule
\multicolumn{1}{l|}{Models} & \multicolumn{1}{c}{Linear}\\
\midrule
\;BoWNet $K=512$  & 69.76\\
\;BoWNet $K=1024$ & 70.43\\
\;BoWNet $K=2048$ & \textbf{71.01}\\
\;BoWNet $K=4096$ & 70.99\\
\bottomrule
\end{tabular}}}
\caption{
\textbf{CIFAR-100 linear classifier results with WRN-28-10. Impact of vocabulary size.}
Here we used an initial version of our method implemented with less aggressive augmentation techniques.}
\vspace{-12pt}
\label{tab:UnsupervisedCIFAR100ResultsBoWImpact}
\end{table}
\begin{table}[t!]
\centering
\renewcommand{\figurename}{Table}
\renewcommand{\captionlabelfont}{\bf}
\renewcommand{\captionfont}{\small} 
{\setlength{\extrarowheight}{2pt}\small
{
\begin{tabular}{ l  | c  c  c  c | c  }
\toprule
\multicolumn{1}{l|}{Method} & \multicolumn{1}{c}{$n=$1} & \multicolumn{1}{c}{5} & \multicolumn{1}{c}{10} & \multicolumn{1}{c|}{50} & \multicolumn{1}{c}{Linear}\\
\midrule
\;RotNet            & 58.3 & 74.8 & 78.3 & 81.9 & 60.3\\
\;BoWNet            & 69.1 & 86.3 & 89.2 & 92.4 & 71.5\\  
\midrule
\multicolumn{2}{l}{\textbf{\;Additional ablations}}\\
\;BoWNet - predict $y(\tilde{\vx})$ & 61.7 & 80.0 & 83.4 & 87.4 & 61.3\\
\;BoWNet - linear $\Omega(\cdot)$   & 70.4 & 85.6 & 88.3 & 90.7 & 63.3\\
\;BoWNet - binary                   & 70.1 & 86.8 & 89.5 & 92.7 & 71.4\\
\bottomrule
\end{tabular}}}
\caption{
\textbf{CIFAR-100 linear classifier and few-shot results.}
For these results we use the WRN-28-10.}
\label{tab:UnsupervisedCIFAR100ResultsAblations}
\end{table}

\section{Additional experimental analysis}

\subsection{CIFAR-100 results} \label{sec:additional_exp_cifar100}
Here we provide an additional ablation analysis of our method on the CIFAR-100 dataset.
As in section~\S\ref{sec:method_analysis}, %
we use the WRN-28-10~\cite{zagoruyko2016wide} architecture.

\textbf{Impact of vocabulary size.}
In Table~\ref{tab:UnsupervisedCIFAR100ResultsBoWImpact} we report linear classification results for different vocabulary sizes $K$. 
We see that increasing the vocabulary size from $K=256$ to $K=2048$ in CIFAR-100, offers obvious performance improvements. 
Then, from $K=2048$ to $K=4096$, there is no additional improvement.

\paragraph{Reparametrized linear layer for $\Omega(\cdot)$.}
In section~\S\ref{sec:predict_bow}, %
we described the linear-plus-softmax prediction layer $\Omega(\cdot)$ implemented with a reparametrized version of the standard linear layer.
In this reparametrized version, 
instead of directly applying the weight vectors $W = [\vw_1, \cdots, \vw_K]$ to a feature vector $\Phi(\tilde{\vx})$,
we first $L_2$-normalize the weight vectors, and then apply a unique learnable magnitude $\gamma$ for all the weight vectors (see equation (5) of main paper).
The goal is to avoid always favoring the most frequently occurring words in the dataset by letting the linear layer of $\Omega(\cdot)$ learn a different magnitude for the weight vector of each word (which is what happens in the case of the standard linear layer). In terms of effect over the weight vector, this reparametrization is similar with the power-law normalization~\cite{perronnin2010improving} used for mitigating the burstiness effect on BoW-like representations~\cite{jegou2009burstiness}. 
Here, we examine the impact of the chosen reparametrization by providing in Table~\ref{tab:UnsupervisedCIFAR100ResultsAblations} results for the case of implementing $\Omega(\cdot)$ with a standard linear layer (entry BoWNet - linear).
We see that with the standard linear layer the performance of the BoWNet model deteriorates, especially on the linear classification metric, 
which validates our design of the $\Omega(\cdot)$ layer.

\paragraph{Predicting $y(\tilde{\vx})$ instead of $y(\vx)$.}
In our work, given a perturbed image $\tilde{\vx}$, we train a convnet to predict the BoW representation $y(\vx)$ of the original image $\vx$.
The purpose of predicting the BoW of the original image $\vx$ instead of the perturbed one $\tilde{\vx}$, is to force the convnet to learn perturbation-invariant and context-aware features.
We examine the impact of this choice by providing in Table~\ref{tab:UnsupervisedCIFAR100ResultsAblations} results for when the convnet is trained to predict the BoW of the perturbed image $\tilde{\vx}$ (entry BoWNet - predict $y(\tilde{\vx})$).
As expected, in this case there is a significant drop in the BoWNet performance. 

\paragraph{Histogram BoW vs Binary BoW}
In section \S\ref{sec:create_bow} %
we describe two ways for reducing the visual word description of an image to a BoW representation.
Those are, \textbf{(1)} to count the number of times each word appears in the image (see equation (\ref{eq:bow_count})), called Histogram BoW,
and \textbf{(2)} to just indicate for each word whether it appears in the image (see equation (\ref{eq:bow_bin}))~\cite{sivic2006video, jegou2009packing}, called Binary BoW.
In Table~\ref{tab:UnsupervisedCIFAR100ResultsAblations} we provide evaluation results with both the histogram version (entry BoWNet) and the binary version (entry BoWNet - binary).
We see that they achieve very similar linear classification and few-shot performance.

\subsection{Small-scale experiments on ImageNet} \label{sec:imagenet_BoW_design}
\begin{table}[t!]
\centering
\renewcommand{\figurename}{Table}
\renewcommand{\captionlabelfont}{\bf}
\renewcommand{\captionfont}{\small} 
{\setlength{\extrarowheight}{2pt}\small
{
\begin{tabular}{ l  | c || l | c }
\toprule
\multicolumn{1}{l|}{BoW from} & \multicolumn{1}{c||}{linear cls} & \multicolumn{1}{l|}{Vocabulary} & \multicolumn{1}{c}{linear cls}\\
\midrule
\;\texttt{conv3} & 42.08 & \;$K=2048$  & 45.38\\
\;\texttt{conv4} & \textbf{45.38} & \;$K=8192$  & 46.03\\
\;\texttt{conv5} & 40.37 & \;$K=20000$ & \textbf{46.45}\\
\midrule
\multicolumn{2}{l||}{\textbf{\;Vocabulary size $K=2048$}} & \multicolumn{2}{l}{\textbf{\;BoW from \texttt{conv4}}}\\
\bottomrule
\end{tabular}}}
\caption{\textbf{ResNet18 small-scale experiments on ImageNet. Linear classifier results.} %
The accuracy of the RotNet model used for building the BoW representations is 37.61.
The left section explores the impact of the layer (of RotNet) that we use for building the BoW representations (with $K=2048$).
The right section explores the impact of the vocabulary size $K$ (with BoW from the \texttt{conv4}).}
\vspace{-8pt}
\label{tab:SmallScaleImageNet}
\end{table}

Here we provide an additional experimental analysis of our method on the ImageNet dataset.
Specifically, we study the impact of the feature block of the base convnet and the vocabulary size that are used for building the BoW representation.
Due to the computationally intensive nature of ImageNet, we analyze those aspects of our method by performing ``small-scale'' ImageNet experiments.
By ``small-scale'' we mean that we use the light-weight ResNet18 architecture and we train using only $20\%$ of ImageNet training images, and for few epochs.

\textbf{Implementation details.}
We train the self-supervised models with SGD for $48$ epochs.
The learning rate is initialized at $0.1$ and dropped by a factor of $10$ after $15$, $30$, and $45$ epochs.
The batch size is $200$ and weight decay $5e-4$.

\textbf{Evaluation protocols.}
We evaluate the learned self-supervised representations by freezing them and then training on top of them $1000$-way linear classifiers for the ImageNet classification task.
The linear classifier is applied on top the feature map of the last residual block of ResNet18, resized to $512 \times 4 \times 4$ with adaptive average pooling.
It is trained with SGD for $60$ epochs using a learning rate of $0.1$ that is dropped by a factor of $10$ every $20$ epochs.
The weight decay is $1e-3$. 

\textbf{Results.}
We report results in Table~\ref{tab:SmallScaleImageNet}.
First, we study the impact on the quality of the learned representations of the RotNet feature block that is used for building the BoW representation.
In the left section of Table~\ref{tab:SmallScaleImageNet} we report results for the cases of (a) \texttt{conv3} (2nd residual block), (b) \texttt{conv4} (3rd residual block), and \texttt{conv5} (4th residual block).
We see that the best performance is for the \texttt{conv4}-based BoW.
Furthermore, in the right section of Table~\ref{tab:SmallScaleImageNet} we examine the impact of the vocabulary size $K$ on the quality of the learned representations.
We see that increasing the vocabulary size from $K=2048$ to $K=20000$ leads to significant improvement for the linear classifier.
In contrast, in Table~\ref{tab:UnsupervisedCIFAR100ResultsBoWImpact} with results on CIFAR-100, we saw that increasing the vocabulary size after $K=2048$ does not improve the quality of the learned representations. 
Therefore, it seems that the optimal vocabulary size depends on the complexity of the dataset to which we apply the BoW prediction task. 

\subsection{Full ImageNet and Places205 classification results} \label{sec:full_imnet_places}
\begin{table*}[t!]
\centering
\renewcommand{\figurename}{Table}
\renewcommand{\captionlabelfont}{\bf}
\renewcommand{\captionfont}{\small} 
{\setlength{\extrarowheight}{2pt}\small
{
\begin{tabular}{ l  | c c c c c | c c c c c }
\toprule
       & \multicolumn{5}{c|}{ImageNet}   & \multicolumn{5}{c}{Places205}\\    
Method & \texttt{conv2} & \texttt{conv3} & \texttt{conv4} & \texttt{conv5} & \texttt{pool5} & \texttt{conv2} & \texttt{conv3} & \texttt{conv4} & \texttt{conv5} & \texttt{pool5}\\
\midrule
\;Random~\cite{goyal2019scaling}    & 13.7 & 12.0 & 8.0 & 5.6   & -    & 16.6 & 15.5 & 11.6 & 9.0 & -\\
\multicolumn{11}{l}{\textbf{\;Supervised methods}}\\
\;ImageNet~\cite{goyal2019scaling}  & 33.3 & 48.7 & 67.9 & 75.5 & -    & 32.6 & 42.1 & 50.8 & 51.5 & -\\
\;ImageNet$^\ast$                   & 32.8 & 47.0 & 67.2 & 76.0 & 76.2 & 35.2 & 42.6 & 50.9 & 52.8 & 52.0\\
\;Places205~\cite{goyal2019scaling} & 31.7 & 46.0 & 58.2 & 51.7 & -    & 32.3 & 43.2 & 54.7 & 62.3 & -\\
\midrule
\multicolumn{11}{l}{\textbf{\;Prior self-supervised methods}}\\
\;RotNet$^\ast$                         & 30.1 & 42.0 & 52.5 & 46.2 & 40.6 & 32.9 & 40.1 & 45.0 & 42.0 & 39.4\\
\;Jigsaw~\cite{goyal2019scaling}        & 28.0 & 39.9 & 45.7 & 34.2 & -    & 28.8 & 36.8 & 41.2 & 34.4 & -\\
\;Colorization~\cite{goyal2019scaling}  & 24.1 & 31.4 & 39.6 & 35.2 & -    & 28.4 & 30.2 & 31.3 & 30.4 & -\\  
\;\textcolor{mygray}{LA$^\dagger$~\cite{zhuang2019local}}  & \textcolor{mygray}{23.3} & \textcolor{mygray}{39.3} & \textcolor{mygray}{49.0} & \textcolor{mygray}{60.2} & - & \textcolor{mygray}{26.4} & \textcolor{mygray}{39.9} & \textcolor{mygray}{47.2} & \textcolor{mygray}{50.1} & -\\
\multicolumn{11}{l}{\textbf{\;Concurrent work}}\\
\;MoCo~\cite{he2019momentum}      & - & - & - & - & \underline{60.6} & - & - & - & - & - \\
\;PIRL~\cite{misra2019self}       & 30.0 & 40.1 & 56.6 & \textbf{63.6} & - & 29.0 & 35.8 & 45.3 & \underline{49.8} & - \\
\;\textcolor{mygray}{CMC$^\ddagger$~\cite{tian2019contrastive}} & - & - & - & - & \textcolor{mygray}{64.1} & - & - & - & - & - \\
\midrule
\;BowNet \texttt{conv4}            & \textbf{34.4} & \underline{48.7} & \underline{60.0} & \underline{62.5} & \textbf{62.1} & 36.7 & \textbf{44.7} & \textbf{50.5} & \textbf{50.9} & \textbf{51.1}\\
\;BowNet \texttt{conv5}           & \underline{34.2} & \textbf{49.1} & \textbf{60.5} & 60.4 & 60.2 & \textbf{36.9} & \textbf{44.7} & \underline{50.1} & 49.6 & \underline{49.5}\\
\bottomrule
\end{tabular}}}
\caption{
\textbf{\resnetfifty top-1 center-crop linear classification accuracy on ImageNet and Places205.} 
For the \texttt{conv2-conv5} layers of \resnetfifty, to evaluate linear classifiers, we resize their feature maps to around $9k$ dimensions (in the same way as in~\cite{goyal2019scaling}).
\texttt{pool5} indicates the accuracy of the linear classifier trained on the $2048$-dimensional feature vectors produced by the global average pooling layer after \texttt{conv5}.
$^\dagger$: LA~\cite{zhuang2019local} uses 10 crops for evaluation.
$^\ddagger$: CMC~\cite{tian2019contrastive} uses two \resnetfifty feature extractor networks instead of just one.
$^\ast$: our implementation.}
\vspace{-8pt}
\label{tab:ImageNetPlacesResults}
\end{table*}

In Table~\ref{tab:ImageNetPlacesResults} we provide the full experimental results of our method on the ImageNet and Places205 classification datasets.

\section{Implementation details}
\subsection{Implementing RotNet} \label{sec:imp_details_rotnet}

For the implementation of the rotation prediction network, RotNet, we follow the description and settings from Gidaris \etal~\cite{gidaris2018unsupervised}. 
RotNet is composed of a feature extractor $\hat{\Phi}(\cdot)$ and a rotation prediction module.
The rotation prediction module gets as input the output feature maps of $\hat{\Phi}(\cdot)$ and is implemented as a convnet.
It consists of a block of residual layers followed by global average pooling and a fully connected classification layer.
In the CIFAR-100 experiments where $\hat{\Phi}(\cdot)$ is implemented with a WRN-28-10 architecture, the residual block of the rotation prediction module has 4 residual layers (similar to the last residual block of WRN-28-10), with $640$ feature channels as input and output.
In the MiniImageNet experiments where $\hat{\Phi}(\cdot)$ is implemented with a WRN-28-4 architecture, the residual block of the rotation prediction module has again 4 residual layers, but with  $256$ feature channels as input and output.
Finally, in ImageNet experiments with ResNet50, the residual block of the rotation prediction module has 1 residual layer with $2048$ feature channels as input and output.

During training for each image of a mini-batch, we generate its four rotated copies ($0^{\circ}$, $90^{\circ}$, $180^{\circ}$, and $270^{\circ}$ $2$D rotations) and predict the rotation class of each copy. For supervision we use the cross-entropy loss over the four rotation classes. 

After training we discard the rotation prediction module and consider only the feature extractor $\hat{\Phi}(\cdot)$ for the next stages, \ie spatially dense descriptions and visual words.

\subsection{Building BoW for self-supervised training} \label{sec:imp_details_imagenet_BoW}

For the ImageNet experiments, 
given an image, we build its target BoW representation using visual words extracted from both the original and the horizontally flipped version of the image.
Also, for faster training we pre-cache the BoW representations.
Finally, in all experiments, when computing the target BoW representation we ignore the visual words that correspond to the feature vectors on the edge of the feature maps.

\subsection{Few-shot protocol} \label{sec:imp_details_fewshot}

The typical pipeline in few-shot learning is to first train a model on a set of \emph{base} classes and then to evaluate it on a different set of \emph{novel} classes (each set of classes is split into train and validation subsets). For MiniImageNet experiments we use this protocol, as this dataset has three different splits of classes: $64$ training classes, $16$ for validation, and $20$ for test. For the few-shot experiments on CIFAR-100 we do not have such splits of classes and we adjust this protocol by selecting a subset of $20$ classes and sample from the corresponding test images for evaluation. In this case, the feature extractor $\Phi(\cdot)$ is trained in a self-supervised manner on train images from all $100$ classes of CIFAR-100.

Few-shot models are evaluated over a large number of few-shot tasks: we consider here $2,000$ tasks. The few-shot evaluation tasks are formed by first sampling $t$ categories from the set of novel/evaluation classes and then selecting randomly $n$ training samples and $m$ test samples per category. The classification performance is measured on the $t \times m$ test images and is averaged over all sampled few-shot tasks. For few-shot experiments we use $t=5$, $m=15$, $n \in \{1,5, 10, 50\}$.

\end{document}